\documentclass{article}

\usepackage{microtype}
\usepackage{graphicx}
\usepackage{subcaption}
\usepackage{booktabs} 
\usepackage{multirow} 
\usepackage{hyperref}
\usepackage{tabularx, booktabs}
\usepackage{tcolorbox}
\usepackage{xcolor}
\newcommand{\wrong}[1]{{\color{red}#1}}
\newcommand{\correct}[1]{{\color{green!60!black}#1}}
\usepackage{tcolorbox}
\tcbuselibrary{skins,breakable}
\usepackage{siunitx}
\usepackage{adjustbox}
\usepackage[table]{xcolor}

\usepackage[preprint]{icml2026}

\usepackage{amsmath}
\usepackage{amssymb}
\usepackage{mathtools}
\usepackage{amsthm}
\usepackage{amsmath} 
\DeclareMathOperator*{\softmax}{softmax}

\usepackage[capitalize,noabbrev]{cleveref}

\theoremstyle{plain}

\theoremstyle{definition}

\theoremstyle{remark}

\usepackage[textsize=tiny]{todonotes}

\icmltitlerunning{Thinking by Subtraction:
Confidence-Driven Contrastive Decoding for LLM Reasoning}

\begin{document}

\twocolumn[
  \icmltitle{Thinking by Subtraction: \\
  Confidence-Driven Contrastive Decoding for LLM Reasoning}



  \icmlsetsymbol{equal}{*}

\begin{icmlauthorlist}
\icmlauthor{Lexiang Tang}{equal,sch}
\icmlauthor{Weihao Gao}{equal,sch}
\icmlauthor{Bingchen Zhao}{sch1}
\icmlauthor{Lu Ma}{sch}
\icmlauthor{Qiao jin}{sch}
\icmlauthor{Bang Yang}{sch}
\icmlauthor{Yuexian Zou}{sch}
\end{icmlauthorlist}

\icmlaffiliation{sch}{Peking University, Beijing, China}
\icmlaffiliation{sch1}{University of Edinburgh, Edinburgh, UK}

\icmlcorrespondingauthor{Bang Yang}{yangbang@pku.edu.cn}
\icmlcorrespondingauthor{Yuexian Zou}{zouyx@pku.edu.cn}

\icmlkeywords{Machine Learning, LLM Reasoning, Contrastive Decoding, ICML}

\vskip 0.3in
] 



\printAffiliationsAndNotice{}  

\begin{abstract}
Recent work on test-time scaling for large language model (LLM) reasoning typically assumes that allocating more inference-time computation uniformly improves correctness.
However, prior studies show that reasoning uncertainty is highly localized: a small subset of low-confidence tokens disproportionately contributes to reasoning errors and unnecessary output expansion.
Motivated by this observation, we propose Thinking by Subtraction, a confidence-driven contrastive decoding approach that improves reasoning reliability through targeted token-level intervention.
Our method, termed Confidence-Driven Contrastive Decoding (CCD), identifies low-confidence tokens (LC-tokens) during decoding and selectively intervenes at these positions.
To construct a contrastive reference, CCD replaces high-confidence tokens (HC-tokens) with semantically minimal placeholder symbols, yielding a deliberately confused contrastive distribution.
By subtracting this reference distribution at LC-token positions, CCD refines uncertain predictions without additional training or multiple reasoning trajectories.
Experimental results demonstrate that CCD consistently improves accuracy across multiple mathematical reasoning benchmarks and achieves higher correctness with substantially more concise outputs on challenging AIME-style tasks. By incurring only minimal key–value (KV) cache overhead, CCD serves as a resource-efficient alternative to uniform test-time scaling. Our findings suggest that targeted intervention at localized low-confidence states can stabilize the decoding process and enhance reasoning reliability without the need for extensive computational redundancy. Our code will be made available at: https://github.com/bolo-web/CCD.
\end{abstract}

\section{Introduction}

\begin{figure}
    \centering
    \includegraphics[width=1.0\linewidth]{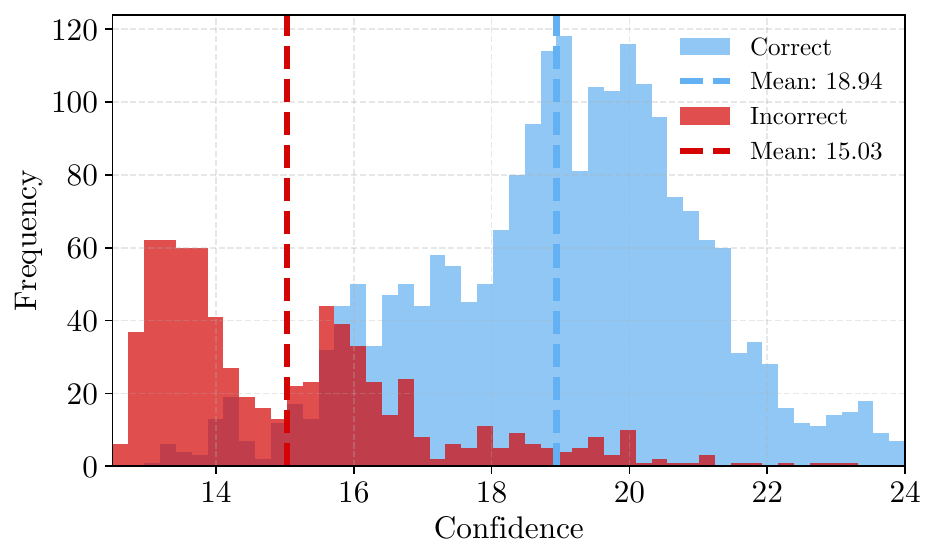}
\caption{
Trajectory-level relationship between token confidence and final answer correctness.
Across 2{,}880 reasoning trajectories generated by Qwen3-8B on AIME24.
}
    \label{fig:intro}
\end{figure}

Recent progress in large language models (LLMs) ~\cite{gpt4,qwen3,gemini,llama,deepseek} has shown that increasing inference-time computation can substantially improve reasoning performance.
However, most test-time scaling strategies~\cite{lightman2023let,chen2024not, welleck2024scaling,wang2022self,zheng2025parallel,zhao2025majority} achieve these gains by uniformly expanding computation across the entire decoding process, often through longer deliberation chains or multiple reasoning trajectories.
While effective, such approaches incur significant computational overhead and tend to amplify unnecessary reasoning.

Motivated by these limitations, and by recent evidence that controlled modification of internal logit and attention distributions can enhance reasoning fidelity in LLMs~\citep{nguyen2024turning,karan2025reasoning,chen2025exploration,topn,visflow,zhao2026gift}, we explore an alternative direction that focuses on improving the quality of a single decoding trajectory, rather than allocating additional computation globally or generating multiple reasoning branches.

Unlike test-time scaling, contrastive decoding (CD) has recently emerged as a lightweight alternative for enhancing reasoning performance at inference time, without requiring additional training or multiple sampling paths~\citep{chuang2023dola}.
Existing CD-based methods improve generation quality by contrasting the main model with auxiliary distributions induced by weakened or perturbed contexts, thereby suppressing undesirable token candidates during decoding. These results highlight the effectiveness of contrastive signals for shaping model behavior at test time.

However, prior CD-based approaches to reasoning~\citep{chuang2023dola,o2023contrastive,leng2024mitigating,phan2024distillation,zhao2024enhancing,UCD} typically apply contrastive signals uniformly across decoding steps, implicitly treating all tokens as equally uncertain.
Related efforts explore coarser-grained interventions, such as layer-wise contrastive control~\citep{chuang2023dola,zhang2025active}, which operate at the representation level rather than targeting localized uncertainty during autoregressive decoding. In practice, autoregressive reasoning uncertainty is often highly localized to specific decision points (e.g., arithmetic operations or logical transitions), while many tokens are generated with high confidence. Consequently, indiscriminate application of contrastive decoding may dilute its impact on critical steps and incur unnecessary computation on already-certain predictions.

Motivated by findings in DeepConf~\citep{fu2025deepconf} that predictive confidence is closely correlated with reasoning correctness, we ask whether contrastive decoding can be used to selectively intervene at low-confidence decoding states and improve reasoning by increasing predictive confidence at these positions.
While this analysis does not establish token-level causality, it indicates that low-confidence decoding states serve as reliable signals of unstable reasoning within a trajectory.



Based on this observation, we hypothesize that selectively improving token reliability at low-confidence decoding positions can mitigate downstream reasoning errors.
This motivates a decoding-time mechanism based on contrastive decoding, which identifies locally uncertain reasoning steps and intervenes only when necessary.


In this work, we introduce a confidence-aware perspective on contrastive decoding that directly addresses this limitation.
Rather than treating CD as a global decoding strategy, we use token-level confidence as an explicit uncertainty signal to determine \emph{when} and \emph{where} contrastive decoding should be activated.
This selective formulation allows CCD to preserve stable reasoning behavior at high-confidence positions while focusing contrastive correction on locally uncertain steps.

The main contributions can be summarized as follows:
\begin{itemize}
    \item We introduce \textbf{Confidence-Driven Contrastive Decoding (CCD)}, a decoding-time framework that unifies token-level confidence estimation with contrastive decoding to correct locally uncertain reasoning steps.
    \item We propose a \textbf{selective intervention strategy} that activates contrastive decoding only at low-confidence chain-of-thought tokens, enabling fine-grained control and avoiding the uniform computational overhead of sampling-based or multi-trajectory methods.
    \item We demonstrate that CCD is \textbf{training-free, model-agnostic, and efficient}, requiring no additional supervision or architectural changes, while introducing minimal and predictable inference-time overhead through bounded KV cache maintenance.
\end{itemize}

In Section~2, we introduce \emph{Confidence-Driven Contrastive Decoding} (CCD).
Section~3 presents empirical and statistical evaluations of CCD on multiple reasoning benchmarks.
Section~4 concludes the paper and discusses limitations and future directions.

\section{Methodology}

\subsection{Overview}
\begin{figure*}
    \centering
    \includegraphics[width=1\linewidth]{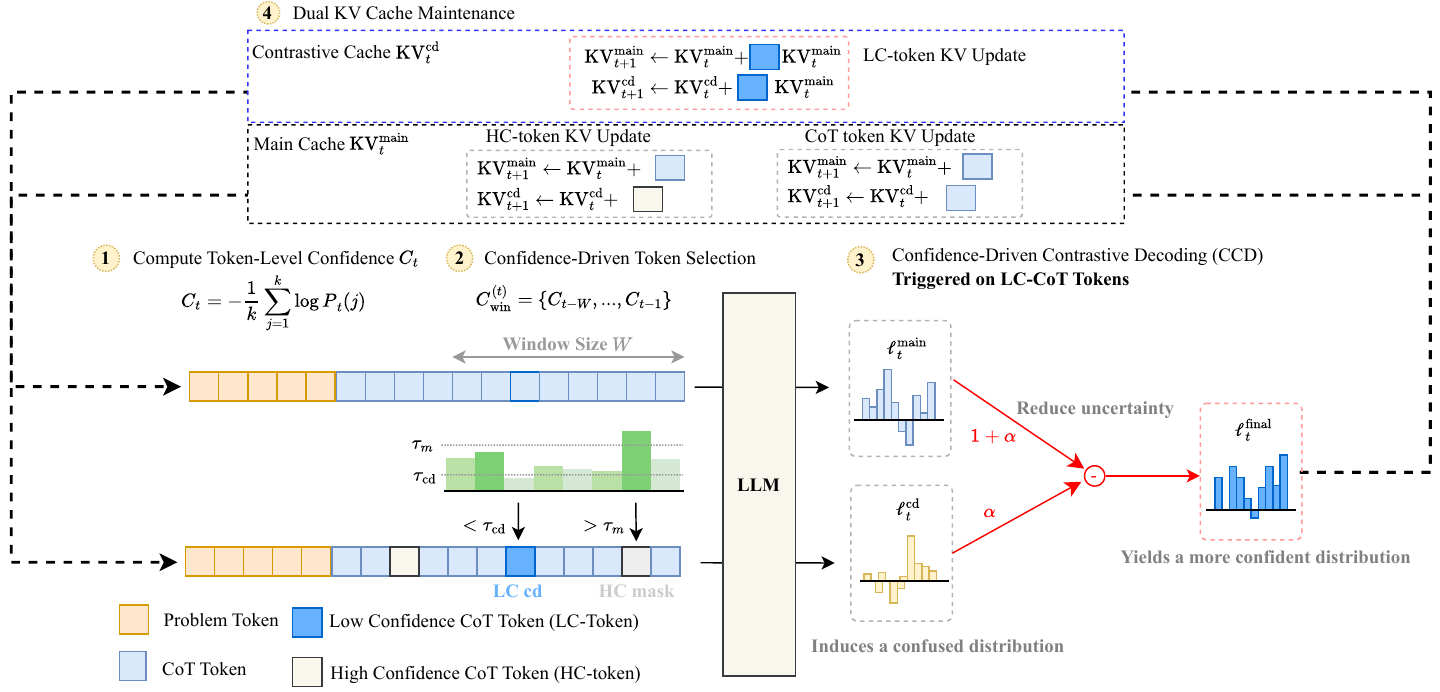}
    \caption{Overview of Confidence-Driven Contrastive Decoding (CCD).
    The decoding process consists of four key components:
    (1) online estimation of token-level confidence;
    (2) confidence-driven token selection that partitions chain-of-thought tokens into low-confidence (LC) and high-confidence (HC) sets;
    (3) contrastive decoding triggered at LC-CoT tokens to refine uncertain predictions;
    and (4) dual key--value (KV) cache maintenance to support selective intervention without disrupting standard autoregressive decoding.}
    \label{fig:placeholder}
\end{figure*}
We propose \emph{Confidence-Driven Contrastive Decoding} (CCD), a decoding-time approach for improving the reliability of autoregressive reasoning in large language models.
CCD is motivated by the observation that reasoning errors tend to arise at locally uncertain decoding steps, where token-level predictive confidence is low and semantic constraints are weakened.
Rather than uniformly increasing inference-time computation, CCD performs targeted intervention only at such uncertain positions.

\subsection{Preliminaries: Confidence-Based Uncertainty and Contrastive Decoding}
\noindent\textbf{Token-Level Confidence.}
CCD leverages token-level predictive confidence as a control signal for identifying unstable decoding states.
Following prior work, the confidence at position $i$ is defined as
\begin{equation}
C_i = - \frac{1}{k} \sum_{j=1}^{k} \log P_i(j),
\end{equation}
where $P_i(j)$ denotes the probability of the $j$-th most likely token and $k$ is the number of top-ranked candidates.
Lower confidence values correspond to flatter predictive distributions and higher uncertainty.

\noindent\textbf{Sliding-Window Confidence.}
Following prior work on sliding-window statistics~\citep{fu2025deepconf}, we maintain a sliding window of recent token-level predictive confidence at decoding step $t$:
\begin{equation}
\label{eq:conf_window}
\mathcal{C}_{\mathrm{win}}^{(t)} = \left\{ C_{t-W}, \dots, C_{t-1} \right\}.
\end{equation}
This window provides a local estimate of confidence dynamics for adaptive decoding control.

\noindent\textbf{Contrastive Decoding.}
To refine predictions at low-confidence steps, CCD employs contrastive decoding to reshape the local decoding distribution.
At decoding step $t$, logits are computed from the original context $c_t$ and a contrastive context $\tilde{c}_t$ as
$\ell^{(1)}_t = \operatorname{logit}_\theta(\cdot \mid c_t)$
and
$\ell^{(2)}_t = \operatorname{logit}_\theta(\cdot \mid \tilde{c}_t)$.
The contrastive logits are then given by
\begin{equation}
\ell^{\mathrm{cd}}_t = (1+\alpha)\,\ell^{(1)}_t - \alpha\,\ell^{(2)}_t,
\end{equation}
with the resulting decoding distribution
\begin{equation}
p_{\mathrm{cd}}\!\left(y_t \mid c_t, \tilde{c}_t\right)
= \operatorname{softmax}\!\left(\ell^{\mathrm{cd}}_t\right).
\end{equation}

Overall, CCD operates entirely at decoding time, requires no additional training or multiple reasoning trajectories, and introduces minimal computational overhead, providing a simple and principled mechanism for uncertainty-aware reasoning during inference.

\noindent\textbf{Confidence-Driven Token Selection.}
CCD performs token-level selection by partitioning chain-of-thought (CoT) tokens into two confidence-aware categories based on percentile statistics of recent decoding states.
Specifically, given the sliding-window confidence set $\mathcal{C}_{\mathrm{win}}^{(t)}$, we estimate adaptive quantile-based thresholds to distinguish \emph{low-confidence} and \emph{high-confidence} tokens.

We use the threshold $\tau_{cd}$ to identify Low-Confidence CoT tokens (\textbf{LC-Tokens}), defined as tokens whose confidence falls below the $q_{cd}$-th quantile of $\mathcal{C}_{\mathrm{win}}^{(t)}$. Such tokens indicate localized uncertainty in the reasoning trajectory and serve as intervention points for contrastive decoding:
\begin{equation}
\mathbb{I}^{\mathrm{cd}}_t =
\mathbf{1}\!\left(C_t < \tau_{cd} \ \wedge\ t \in \mathcal{R}\right).
\end{equation}
Here, $\mathcal{R}$ denotes the set of decoding steps corresponding to the model-generated reasoning region, excluding question tokens.

Using the threshold $\tau_{rep}$, we identify High-Confidence CoT tokens (\textbf{HC-Tokens}) as those above the $q_r$-th quantile of $\mathcal{C}_{\mathrm{win}}^{(t)}$. These tokens are considered semantically dominant and are selectively attenuated in the contrastive branch to prevent overconfident cues from dominating the refinement process:
\begin{equation}
\mathbb{I}^{\mathrm{rep}}_t =
\mathbf{1}\!\left(C_t > \tau_{rep} \ \wedge\ t \in \mathcal{R}\right).
\end{equation}

By deriving both LC- and HC-Tokens from local confidence percentiles, CCD dynamically adapts token selection to the evolving uncertainty profile of the decoding trajectory, avoiding reliance on fixed global thresholds.

\subsection{Confidence-Driven Contrastive Decoding}

CCD performs contrastive decoding selectively based on token-level predictive confidence.
The core principle is to intervene only at locally uncertain decoding steps, while preserving standard decoding behavior at confident steps.

\paragraph{Contrastive Distribution Construction.}
At decoding step $t$, when the predictive confidence $C_t$ falls below a threshold, we construct a contrastive decoding distribution by masking high-confidence semantic anchors in the recent context.
Selected tokens are replaced with \emph{Placeholder Tokens} (e.g., \texttt{<endoftext>} or \texttt{<vision\_pad>}), which preserve sequence structure while removing dominant semantic content, yielding a contrastive context that disrupts semantic anchoring without altering positional alignment.
Formally, let $\tilde{\mathbf{y}}_{<t}$ denote the contrastive context obtained from $\mathbf{y}_{<t}$ via placeholder substitution, and the corresponding contrastive logits are computed as
\begin{equation}
\label{eq:cd_logits}
\ell_t^{\mathrm{cd}} = f_\theta(\tilde{\mathbf{y}}_{<t}),
\end{equation}
which induces the contrastive decoding distribution.

\paragraph{Confidence-Driven Logit Subtraction.}
The final logits used for token selection are defined as
\begin{equation}
\label{eq:final_logits}
\ell_t^{\mathrm{final}} =
\begin{cases}
(1+\alpha)\,\ell_t^{\mathrm{main}} - \alpha\,\ell_t^{\mathrm{cd}},
& \text{if } C_t < \tau_{cd} \ \wedge\ t \in \mathcal{R}, \\[0.5em]
\ell_t^{\mathrm{main}},
& \text{otherwise},
\end{cases}
\end{equation}
where $\ell_t^{\mathrm{main}}$ denotes the logits produced from the original decoding context.
The resulting decoding distribution is given by
\begin{equation}
p_{\mathrm{ccd}}(y_t) = \softmax\!\left(\ell_t^{\mathrm{final}}\right),
\end{equation}
from which standard decoding strategies are applied.

\subsection{Theoretical Analysis}

We now explain how CCD constructs an effective contrastive context, and why replacing
high-confidence tokens is a principled choice.

\paragraph{From Confidence Correction to Contrastive Distribution Construction.}
Reasoning errors are concentrated at low-confidence decoding steps, where the predictive
distribution is flat and logit margins are small.
Since contrastive decoding sharpens the main distribution only when the contrastive distribution
is less confident (i.e., higher entropy), the key question becomes how to construct a contrastive
context that yields a deliberately confused reference distribution.

CCD addresses this by selectively masking high-confidence tokens in the decoding history.
Let $c_t=(y_1,\ldots,y_{t-1})$ denote the original context and
$p_\theta(\cdot\mid c_t)=\softmax(\ell_\theta(c_t))$ the corresponding next-token distribution.
The contrastive context $\tilde c_t$ is obtained by replacing high-confidence tokens with a
semantically neutral placeholder $\tilde y$, yielding logits
$\ell_t^{\mathrm{cd}}=\ell_\theta(\tilde c_t)$.

\paragraph{High-Confidence Tokens as Semantic Anchors.}
Let $z_t$ denote the hidden representation used to predict token $y_t$, with logits given by
$\ell = W z_t + b$.
In a transformer decoder, $z_t$ is obtained by contextualizing the embeddings of all previously generated tokens:
\begin{equation}
z_t = F_\theta\!\big(E(y_1), \ldots, E(y_{t-1})\big),
\end{equation}
where $E(\cdot)$ denotes the token embedding function and $F_\theta(\cdot)$ represents the transformer with parameters $\theta$.

Tokens with high predictive confidence typically correspond to sharply peaked conditional
distributions and receive large attention weights at future steps.
Denoting the attention weight on token $y_i$ by
\begin{equation}
a_{t,i}=\softmax\!\left(\frac{q_t^\top k_i}{\sqrt{d}}\right),
\end{equation}
high-confidence tokens often satisfy $a_{t,i}\gg a_{t,j}$ for most $j\neq i$.
Such tokens act as \emph{semantic anchors}, strongly constraining the mapping from context to logits.

\paragraph{Masking Anchors Increases Uncertainty.}
Masking a high-confidence token replaces its embedding $e_i=E(y_i)$ with a placeholder
$\tilde e_i=E(\tilde y)$, inducing a perturbation $\Delta e_i=\tilde e_i-e_i$.
A first-order expansion yields
\begin{equation}
z_t(\tilde c_t)-z_t(c_t)\;\approx\;
J_{t,i}\,\Delta e_i,
\qquad
J_{t,i}=\frac{\partial z_t}{\partial e_i}.
\end{equation}
For anchor tokens, $J_{t,i}$ has large norm because attention routes information through $i$,
leading to a substantial change in $z_t$ and hence in the output logits.
Removing anchor information reduces the mutual information between the context and the next token,
\begin{equation}
I(Y_t;\tilde C_t)\le I(Y_t;C_t),
\end{equation}
which equivalently implies higher conditional entropy,
\begin{equation}
H(Y_t\mid \tilde C_t)\ge H(Y_t\mid C_t).
\end{equation}
As a result, the top-logit margin
$\Delta_t=\ell_t^{(1)}-\ell_t^{(2)}$
is reduced in expectation under contrastive decoding,
i.e., $\Delta_t^{\mathrm{cd}}<\Delta_t^{\mathrm{main}}$,
leading to a flatter and more uncertain reference distribution.

By masking high-confidence tokens when constructing the contrastive reference, we deliberately induce a more confused reference distribution that reflects disrupted semantic anchoring.
Subtracting this distribution during contrastive decoding sharpens predictions at low-confidence positions, thereby increasing token-level confidence and stabilizing the reasoning trajectory.

\subsection{Dual KV Cache Maintenance}

To support contrastive decoding while preserving the original generation trajectory, CCD maintains two parallel key--value (KV) cache states: a main cache $\mathrm{KV}_t^{\mathrm{main}}$ and a contrastive cache $\mathrm{KV}_t^{\mathrm{cd}}$.
The main cache follows standard autoregressive decoding, whereas the contrastive cache encodes a deliberately weakened contextual state used solely for generating contrastive predictions.

During decoding steps within the reasoning phase $\mathcal{R}$, the contrastive cache selectively attenuates locally confident tokens by replacing them with a semantically neutral placeholder.
Formally, the contrastive KV update is defined as
\begin{equation}
\mathrm{KV}_t^{\mathrm{cd}} =
\begin{cases}
\mathrm{KV}(\tilde{y}), 
& C_t > \tau_{rep} \ \wedge\ t \in \mathcal{R}, \\[0.3em]
\mathrm{KV}(y_t), 
& \text{otherwise},
\end{cases}
\end{equation}

where $\tilde{y}$ denotes a \emph{Placeholder Token} as defined above.
Replacing selected tokens with placeholders removes dominant semantic content in the contrastive branch without modifying the decoding procedure.
As a result, CCD introduces only limited overhead by maintaining a single additional contrastive branch with its own KV cache of the same size as standard decoding, while all contrastive operations are performed at the logit level without extra attention passes or model evaluations.


\subsection{Algorithms}

Algorithm~\ref{alg:ccd} presents the pseudocode of our proposed Confidence-Driven Contrastive Decoding (CCD) procedure, providing a complete step-by-step description of the decoding process.

\begin{algorithm}[t]
\caption{Confidence-Driven Contrastive Decoding (CCD)}
\label{alg:ccd}
\begin{algorithmic}

\STATE \textbf{Input:} model $f_\theta$; prompt $y_{1:t_0}$; controlled region $\mathcal{R}$
\STATE \textbf{Input:} window size $W$; quantiles $q_r, q_{cd}$; contrastive weight $\alpha$
\STATE \textbf{Input:} placeholder token $\tilde{y}$; decoder $\mathrm{Dec}(\cdot)$
\STATE \textbf{Output:} generated tokens $y_{t_0+1:T}$

\STATE Initialize main KV cache $\mathrm{KV}^{\mathrm{main}}\gets \mathrm{InitKV}(y_{1:t_0})$
\STATE Initialize contrastive KV cache $\mathrm{KV}^{\mathrm{cd}}\gets \mathrm{InitKV}(y_{1:t_0})$
\STATE Initialize confidence buffer $\mathcal{C}$ with capacity $W$

\FOR{$t=t_0+1$ \textbf{to} $T$}

    \STATE Compute main logits $\ell_t^{\mathrm{main}}\gets f_\theta(\cdot\mid \mathrm{KV}^{\mathrm{main}})$
    \STATE Compute token confidence $C_t$ from $\ell_t^{\mathrm{main}}$ and update $\mathcal{C}$

    \IF{$|\mathcal{C}| = W$}
        \STATE Compute thresholds $\tau_{rep}\gets \mathrm{Quantile}(\mathcal{C}, q_r)$,
               $\tau_{cd}\gets \mathrm{Quantile}(\mathcal{C}, q_{cd})$
    \ELSE
        \STATE $\tau_{rep}\gets +\infty$, $\tau_{cd}\gets -\infty$
    \ENDIF

    \STATE $\mathbb{I}_t^{\mathrm{rep}}\gets \mathbf{1}(t\in\mathcal{R}\wedge C_t>\tau_{rep})$
    \STATE $\mathbb{I}_t^{\mathrm{cd}}\gets \mathbf{1}(t\in\mathcal{R}\wedge C_t<\tau_{cd})$

    \STATE Update contrastive KV cache with
    \[
    x_{t-1}^{\mathrm{cd}} =
    \begin{cases}
    \tilde{y}, & \mathbb{I}_t^{\mathrm{rep}}=1,\\
    y_{t-1}, & \text{otherwise},
    \end{cases}
    \]
    \STATE $\mathrm{KV}^{\mathrm{cd}}\gets \mathrm{KV}^{\mathrm{cd}}\oplus x_{t-1}^{\mathrm{cd}}$

    \IF{$\mathbb{I}_t^{\mathrm{cd}}=1$}
        \STATE Compute contrastive logits $\ell_t^{\mathrm{cd}}\gets f_\theta(\cdot\mid \mathrm{KV}^{\mathrm{cd}})$
        \STATE Fuse logits $\ell_t^{\mathrm{final}}\gets (1+\alpha)\ell_t^{\mathrm{main}}-\alpha\ell_t^{\mathrm{cd}}$
    \ELSE
        \STATE $\ell_t^{\mathrm{final}}\gets \ell_t^{\mathrm{main}}$
    \ENDIF

    \STATE Sample next token $y_t\gets \mathrm{Dec}(\softmax(\ell_t^{\mathrm{final}}))$
    \STATE Update main KV cache $\mathrm{KV}^{\mathrm{main}}\gets \mathrm{KV}^{\mathrm{main}}\oplus y_t$

\ENDFOR

\end{algorithmic}
\end{algorithm}

\section{Empirical and Statistical Evaluation}
\label{sec:evaluation}

\subsection{Experimental Setup}

\subsubsection{Benchmarks}
We conduct experiments on four competitive mathematical reasoning benchmarks:
AIME 2024~\citep{aops2024aime1,aops2024aime2},
AIME 2025~\citep{aops2025aime1,aops2025aime2},
BRUMO 2025~\citep{brumo2025},
and HMMT 2025~\citep{hmmt2025}.
These benchmarks consist of Olympiad-style problems that require multi-step symbolic manipulation and precise logical reasoning.
They have been widely adopted to assess the reasoning capabilities of recent large language models, including Qwen3~\citep{yang2025qwen3}, Grok-4~\citep{xai2025grok4}, and GPT-5~\citep{openai2025gpt5}.
\subsubsection{Models}
We evaluate \emph{Confidence-Driven Contrastive Decoding} (CCD) on a range of recent open-weight large language models spanning diverse parameter scales.
Specifically, we consider Qwen3-4B, Qwen3-8B, Qwen3-14B and Qwen3-32B from the Qwen3 family~\cite{qwen3_4b,qwen3_8b,qwen3_14b,qwen3_moe}, as well as DeepSeek-R1-Distill-Qwen-1.5B~\cite{deepseek_r1_distill_qwen_1_5b}.
All models are evaluated in their original pretrained or distilled forms, without any additional fine-tuning.
\subsubsection{Implementation Details}

\paragraph{Decoding Baselines.}
We compare CCD against standard top-$p$ sampling with $p{=}0.95$ and temperature-based sampling.
Unless otherwise specified, baseline decoding uses temperature sampling with the temperature set to $0.6$.
All decoding methods generate a single reasoning trajectory per input, and no test-time ensembling or self-consistency is applied.
For MTI~\cite{yang2025less}, we follow the decoding configuration reported in the original paper.

\subsection{Main Results}
\subsubsection{Overall Performance}

Table~\ref{tab:main_results} reports the main results of CCD on four competitive mathematical reasoning benchmarks.
Across most datasets and model scales, CCD consistently outperforms the Base decoding strategy and achieves performance comparable to MTI.
In particular, CCD yields stable improvements on both AIME-style benchmarks and more diverse reasoning sets such as BRUMO25 and HMMT25.
On Qwen dense models ranging from 4B to 14B parameters, CCD improves Mean8 accuracy by approximately 3--5\% over the Base method, demonstrating consistent gains across model scales.
Overall, these results indicate that CCD provides a stable improvement over standard decoding under a unified confidence-driven framework.

Table~\ref{tab:aime25_final} further reports the gains of CCD over the Base setting across both mixture-of-experts and dense model variants.
Consistent improvements are observed for MoE models as well as smaller dense models such as DeepSeek-R1-1.5B, suggesting that CCD generalizes well across architectures and model sizes.

\begin{table*}[t]
\centering
\small
\renewcommand{\arraystretch}{1.25}

\begin{tabular*}{\textwidth}{@{\extracolsep{\fill}} l l c c c c}
\toprule
\textbf{Model Series} & \textbf{Method} & \textbf{AIME24} & \textbf{AIME25} & \textbf{BRUMO25} & \textbf{HMMT25} \\
\midrule

\multicolumn{6}{l}{\textit{Large-scale Dense Reasoning Models (Qwen3 Series)}} \\
\midrule

\multirow{3}{*}{Qwen3-4B}
  & Base              & 73.33 & 63.75 & 61.67 & 42.92 \\
  & MTI               & 72.38 & 66.81 & \textbf{67.08} & 40.00 \\
  & \textbf{CCD (Ours)} & \textbf{77.08} & \textbf{67.92} & 65.00 & \textbf{46.25} \\
\midrule

\multirow{3}{*}{Qwen3-8B}
  & Base              & 73.75 & 67.91 & 69.58 & 44.58 \\
  & MTI               & 76.67 & 69.46 & 70.00 & \textbf{53.33} \\
  & \textbf{CCD (Ours)} & \textbf{78.75} & \textbf{71.67} & \textbf{72.50} & 47.92 \\
\midrule

\multirow{3}{*}{Qwen3-14B}
  & Base              & 79.58 & 68.75 & 72.84 & 49.17 \\
  & MTI               & 81.67 & \textbf{73.95} & 73.64 & 50.21 \\
  & \textbf{CCD (Ours)} & \textbf{81.67} & 73.75 & \textbf{75.00} & \textbf{50.83} \\
\bottomrule
\end{tabular*}

\caption{
  Performance evaluation on four mathematical reasoning benchmarks using Qwen3-Think.
  We compare Base (standard decoding), MTI and Ours (CCD) across 8 sampling runs (reported as Mean@8 accuracy \%). 
  \textbf{Bold} indicates the best results, and 
  Results demonstrate that CCD consistently enhances reasoning accuracy by strategically applying contrastive decoding based on confidence thresholds.
}


\label{tab:main_results}
\end{table*}


\begin{table}[t]
\centering
\small
\renewcommand{\arraystretch}{1.3} 
\begin{tabularx}{\columnwidth}{@{\extracolsep{\fill}} l l c c @{}}
\toprule
\textbf{Architecture} & \textbf{Method} & \textbf{AIME25} & \textbf{$\Delta$} \\
\midrule
\multicolumn{4}{l}{\textit{MoE Model: Qwen3-30B-A3B-Thinking-2507 (c2 m5)}} \\
 & Base & 82.08 & -- \\
 & \textbf{CCD (Ours)} & \textbf{85.00} & \textbf{+2.92} \\
\midrule
\multicolumn{4}{l}{\textit{Distilled Model: DeepSeek-R1-Distill-Qwen-1.5B (c3 m5)} } \\
 & Base & 22.08 & -- \\
 & \textbf{CCD (Ours)} & \textbf{24.57} & \textbf{+2.49} \\
\bottomrule
\end{tabularx}
\caption{Performance comparison between Base and CCD on AIME2025 across MoE and Distilled architectures.}
\label{tab:aime25_final}
\end{table}

\subsection{Ablation Studies}

\paragraph{Robustness Across Datasets.}
Due to computational constraints, we evaluate hyperparameter robustness on four representative benchmarks: AIME24, AIME25, BRUMO25, and HMMT25 (Table~\ref{tab:main_results}).
We fix the contrastive decoding threshold $c$ and masking threshold $m$ to their optimal values on AIME25 and directly apply the same configuration to all other benchmarks without additional tuning.
Across datasets with diverse difficulty levels and reasoning characteristics, CCD consistently improves over the Base method and remains competitive with MTI, indicating that its effectiveness does not rely on dataset-specific hyperparameter tuning.

\paragraph{Robustness of the Thresholds $\tau_{cd}$ and $\tau_{rep}$ Across Model Scales}

We evaluate CCD under four representative hyperparameter settings, including $(\tau_{cd},\tau_{rep})=(2,5),(3,5),(3,8),(3,10)$.
As shown in Table~\ref{tab:ablation_cm}, all evaluated configurations consistently outperform the Base decoding across different model scales on AIME25.
These results indicate that the performance gains of CCD are robust to hyperparameter choices and do not rely on a specific setting.

\begin{figure}[t]
    \centering
    \begin{minipage}[t]{0.478\linewidth}
        \centering
        \includegraphics[width=\linewidth]{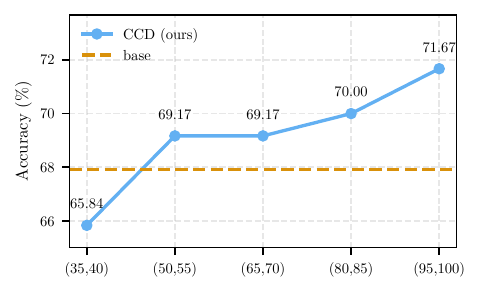}
        \vspace{-0.5em}
    \subcaption{Ablation on different replacement intervals}
        \label{fig:mask_conf_levels}
    \end{minipage}
    \hfill
    \begin{minipage}[t]{0.5\linewidth}
        \centering
        \includegraphics[width=\linewidth]{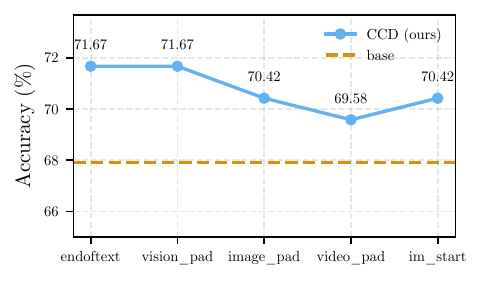}
        \vspace{-0.5em}
        \subcaption{Ablation on using different special token.}
        \label{fig:mask_ablation}
    \end{minipage}
    \vspace{-0.8em}
    \caption{Effects of token masking strategies in the contrastive branch.
    \textbf{(a)} Masking high-confidence tokens induces a more uncertain (higher-entropy) contrastive distribution than masking mid- or low-confidence tokens.
    \textbf{(b)} Ablation on placeholder choices for replacement, where semantically minimal symbols provide the most effective contrastive signal.
}
    \label{fig:mask_combined}
\end{figure}

\paragraph{Robustness of Token Replacement Strategies.}
We conduct ablation studies to evaluate the robustness of the token replacement strategies used to construct the contrastive reference distribution.
Figure~\ref{fig:mask_conf_levels} examines the effect of varying replacement intervals.
Replacing high-confidence tokens consistently induces a more uncertain (higher-entropy) contrastive distribution compared to replacing mid- or low-confidence tokens.
Figure~\ref{fig:mask_ablation} further studies different placeholder choices, showing that semantically minimal symbols yield the most stable and effective contrastive signal.
These results demonstrate that the proposed replacement strategy is robust to threshold selection and provides a principled foundation for confidence-driven contrastive decoding.

\begin{table}[t]
\centering
\begin{tabularx}{\columnwidth}{@{\extracolsep{\fill}} l ccc @{}}
\toprule
& \multicolumn{3}{c}{\textbf{Qwen3 Series (Dense)}} \\
\cmidrule(lr){2-4}
\textbf{$(\tau_{cd},\tau_{rep})$} & \textbf{4B} & \textbf{8B} & \textbf{14B} \\
\midrule
Base   & 63.75 & 67.91 & 68.75 \\
\midrule
(2,5)  & 64.08 & 67.91 & 72.92 \\
(3,5)  & 65.00 & \textbf{71.67} & 71.25 \\
(3,8)  & 67.08 & 70.83 & 73.51 \\
(3,10) & \textbf{67.92} & 68.33 & \textbf{73.75} \\
\bottomrule
\end{tabularx}
\caption{Hyperparameter ablation of $(\tau_{cd},\tau_{rep})$ on AIME25. Base denotes standard decoding without CCD. We evaluate the transferability of the optimal setting across dense architectures.}
\label{tab:ablation_cm}
\end{table}

\subsection{Empirical Analysis}

\begin{figure} 
\centering 
\includegraphics[width=1\linewidth]{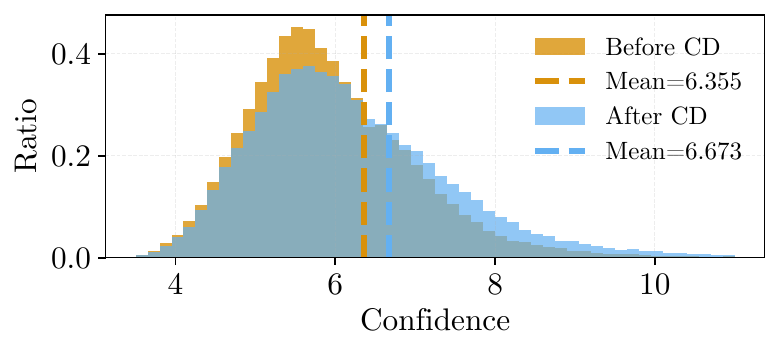} 
\caption{Token-level confidence improvement at low-confidence decoding positions.}\label{fig:token_conf_ana}  \end{figure}

\paragraph{Effect of CCD on Token-Level Confidence.}
We first examine the effect of CCD on token-level confidence at decoding positions identified as low-confidence in the original decoding branch.
As shown in Figure~\ref{fig:token_conf_ana}, CCD consistently increases token-level confidence at these positions after intervention.
Quantitatively, the mean confidence of low-confidence tokens in the original branch increases from $6.355$ to $6.673$ after applying CCD.
This improvement indicates that confidence-driven intervention selectively stabilizes previously uncertain decoding steps, rather than uniformly inflating confidence across all tokens.

\begin{figure} 
\centering 
\includegraphics[width=1\linewidth]{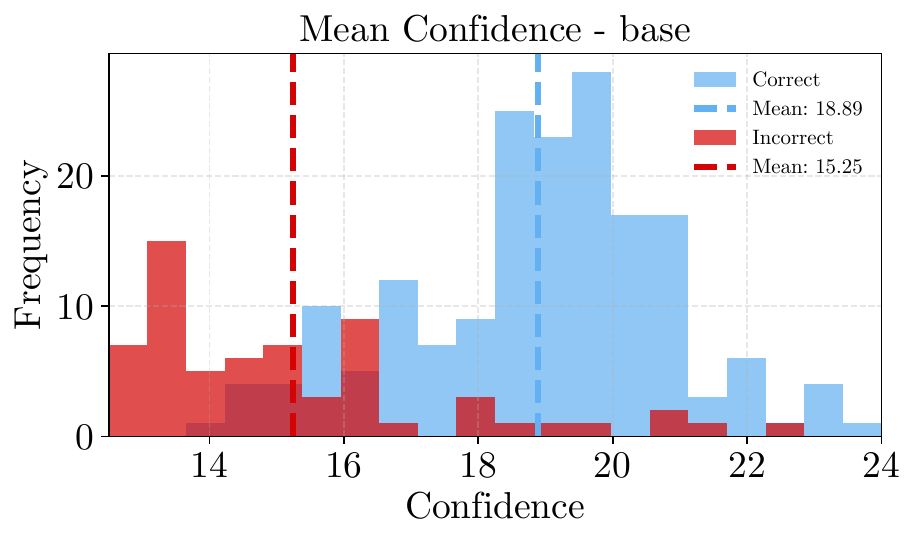} 
    \caption{Trajectory-level confidence comparison before CCD intervention.}
    \label{fig:trace_conf_base}
    \end{figure}

\begin{figure} 
\centering 
\includegraphics[width=1\linewidth]{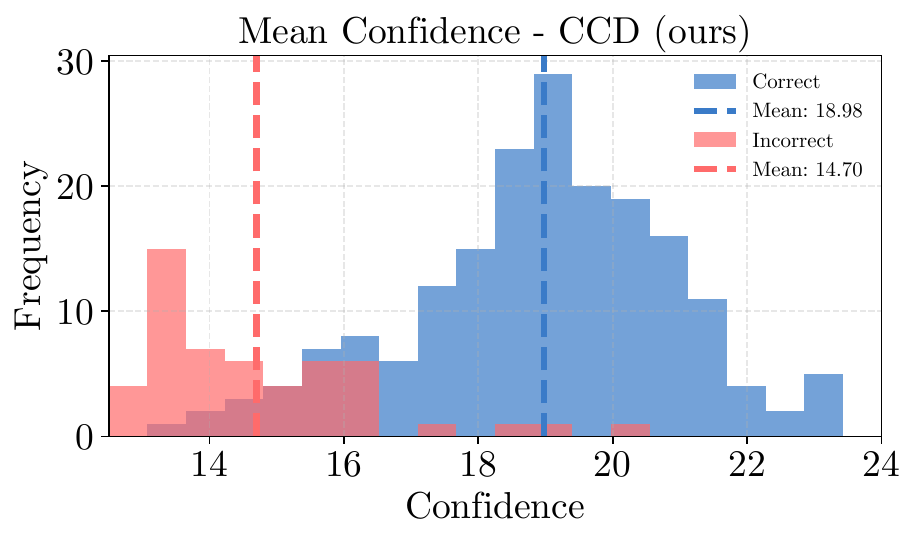} 
    \caption{Trajectory-level confidence comparison after CCD intervention.} 
    \label{fig:trace_conf_ccd} 
    \end{figure}

\paragraph{Effect of CCD on Trace-Level Confidence.}
We further analyze how token-level confidence improvements propagate to the trajectory level.
Figures~\ref{fig:trace_conf_base} and~\ref{fig:trace_conf_ccd} visualize the confidence distributions of complete decoding traces before and after CCD, respectively.
Under the base decoding strategy, correct and incorrect trajectories exhibit substantial overlap, limiting the discriminative power of confidence scores.
After applying CCD, the separability between the two distributions is noticeably improved.
Specifically, the mean confidence of correct trajectories increases from $18.89$ to $18.98$, while the mean confidence of incorrect trajectories decreases from $15.25$ to $14.70$.
These results demonstrate that CCD not only stabilizes local token-level uncertainty, but also enhances trace-level confidence separation, yielding a more reliable confidence signal for downstream verification and reranking.

\paragraph{Effect of CCD on Reasoning Length and Efficiency}

\begin{figure}[t]
    \centering
    \includegraphics[width=1\linewidth]{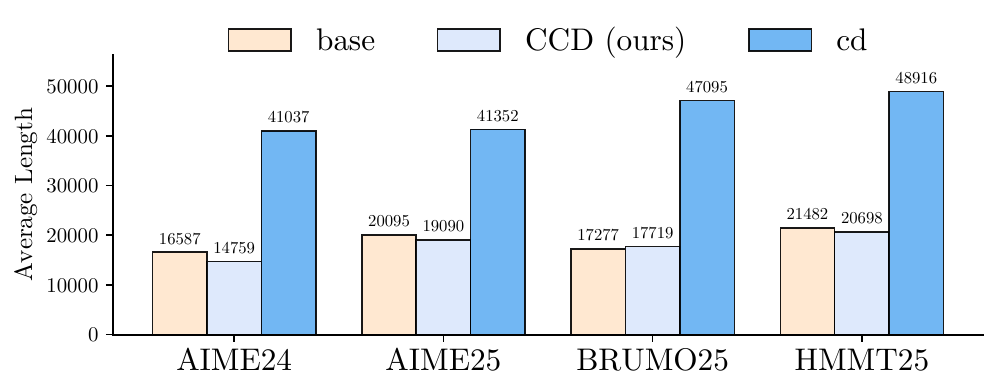}
    \caption{
Average response length of the base output, CCD output, and contrastive distribution for Qwen3-8B across benchmarks.
CCD substantially shortens responses on AIME-style tasks while preserving response length on HMMT25 and BRUMO25, whereas the contrastive distribution yields consistently longer outputs.
}
    \label{fig:len_compare}
\end{figure}

\begin{figure}[t]
    \centering
    \includegraphics[width=1\linewidth]{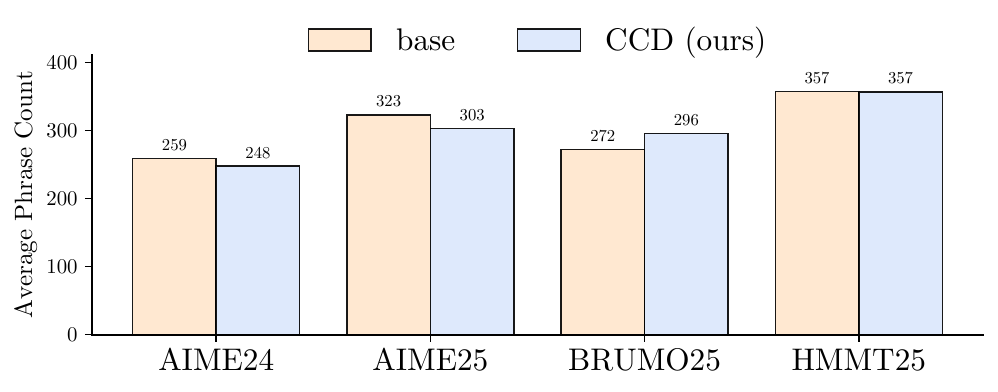}
    \caption{
Frequency of explicit thinking tokens in model responses for Qwen3-8B.
CCD significantly reduces thinking-token usage on AIME benchmarks while leaving HMMT25 and BRUMO25 largely unchanged, indicating selective suppression of overextended reasoning.
}
    \label{fig:think_compare}
\end{figure}

We analyze the effect of CCD on reasoning length and efficiency by comparing the response length and the frequency of explicit thinking tokens across datasets.
Figure~\ref{fig:len_compare} reports the average response length for the base distribution, the final CCD output, and the contrastive distribution.

On AIME24 and AIME25, CCD produces substantially shorter responses than the base method.
This reduction indicates that CCD encourages more compact reasoning trajectories, effectively removing redundant or unnecessary intermediate steps.
In contrast, the responses generated from the contrastive distribution are significantly longer, reflecting the higher uncertainty and weakened semantic constraints induced by masking.
This clear separation confirms that CCD sharpens the final prediction by subtracting a deliberately confused reference distribution rather than truncating reasoning arbitrarily.

On HMMT25 and BRUMO25, the response lengths of the base and CCD outputs largely overlap.
This behavior suggests that CCD does not disrupt stable reasoning patterns when the model is already confident, preserving the original reasoning structure on datasets that do not exhibit strong overthinking.

Figure~\ref{fig:think_compare} further analyzes the frequency of explicit thinking tokens.
On AIME-style benchmarks, CCD significantly reduces the occurrence of such tokens, especially on AIME25, indicating a reduction in repetitive or unproductive deliberation.
In contrast, thinking-token frequency remains nearly unchanged on HMMT25 and BRUMO25, demonstrating that CCD maintains strong stability and avoids unnecessary intervention.
Together, these results show that CCD improves reasoning efficiency by selectively compressing overextended reasoning while leaving confident reasoning trajectories intact.

\section{Conclusion and Limitations}

In this work, we investigate how token-level predictive confidence can be exploited as a decoding-time control signal for improving large language model reasoning.
We propose \emph{Confidence-Driven Contrastive Decoding} (CCD), a simple yet principled approach that performs selective contrastive subtraction at locally unstable decoding states.
Extensive experiments on multiple competitive mathematical reasoning benchmarks demonstrate that CCD consistently improves reasoning accuracy across diverse model scales, without requiring additional training or multiple inference trajectories.
These results suggest that uncertainty-aware decoding offers an efficient alternative to conventional test-time scaling strategies based on extensive sampling or trajectory expansion, achieving a favorable efficiency--performance trade-off with negligible overhead.

Despite these promising results, our study has several limitations.
Due to computational constraints, our evaluation primarily focuses on mathematical reasoning benchmarks, which, while standard for assessing structured multi-step reasoning, do not fully capture the diversity of real-world reasoning tasks.
The effectiveness of CCD on more general-domain benchmarks therefore remains to be systematically explored.
In addition, our experiments are limited to text-only large language models.
Extending CCD to multimodal reasoning settings, such as models jointly processing text and vision, is a natural and promising direction for future work, where token-level confidence may interact with modality-specific representations in more complex ways.
We hope this work encourages further investigation into confidence-guided decoding strategies as a lightweight and broadly applicable approach to enhancing LLM reasoning reliability.

\section*{Impact Statement}

This work proposes Confidence-Driven Contrastive Decoding (CCD) to improve the reliability of large language models in reasoning tasks by mitigating erroneous or low-confidence token generation. While CCD aims to enhance model robustness and reduce reasoning failures, it may still inherit biases present in the underlying pre-trained models and datasets, potentially leading to uneven performance across domains or problem types. Moreover, as with other advanced reasoning techniques, improved reasoning capability could be misused to generate more convincing but incorrect or misleading content if deployed irresponsibly. From a societal perspective, CCD has the potential to improve the dependability of LLM-based systems in high-stakes applications such as education, scientific assistance, and decision support, where reducing unreliable reasoning is particularly important.


\bibliography{example_paper}
\bibliographystyle{icml2026}

\newpage
\appendix
\onecolumn

\section{Related Works}

\subsection{Test-Time Scaling}

Test-time scaling (TTS) enhances LLM reasoning by increasing computational budget during inference rather than training. A prominent line of work focuses on \emph{parallel scaling}, which generates multiple reasoning trajectories and aggregates them through majority voting or reward models~\cite{wang2022self}. However, these methods often suffer from diminishing returns as the number of samples increases, and incur prohibitive computational overhead due to sampling numerous complete reasoning paths~\cite{lightman2023let, chen2024not}.

Recent efforts \cite{relift, li2025leash} enhance inference efficiency via training-based interventions that reshape decoding behavior, leading to shorter reasoning trajectories and reduced autoregressive length. DeepConf~\cite{fu2025deepconf} employs model-internal confidence signals to dynamically filter low-quality reasoning traces during or after generation, achieving significant token reduction while maintaining accuracy. 
Other approaches, such as adaptive consistency methods~\cite{ammar2024era, taubenfeld2025confidence}, aim to determine the optimal number of reasoning chains without exhaustive sampling. 

Unlike these methods that inherently require multiple inference passes to scale performance, our approach demonstrates that effective test-time improvement can be achieved through \emph{single-pass} selective intervention on critical tokens, eliminating the need for multiple reasoning trajectories entirely. This distinguishes our method from both parallel scaling (which samples multiple trajectories) and sequential scaling (which iteratively refines a single trajectory through self-correction~\cite{madaan2023self}).

\subsection{Uncertainty Estimation in LLMs}

Uncertainty estimation has become an important tool for improving the reliability of large language model (LLM) generation.
Existing approaches can be broadly categorized into \emph{entropy-based} and \emph{confidence-based} methods.

\textbf{Entropy-based methods} identify uncertain positions using token-level entropy.
Prior studies~\citep{yang2025less,gatebranch} demonstrate that reasoning uncertainty is highly localized, with a small subset of high-entropy tokens disproportionately influencing output correctness, motivating selective intervention at these positions.

\textbf{Confidence-based methods} leverage the model's predictive confidence as a direct indicator of uncertainty to optimize both model learning and execution. On the learning side, \citet{li2025confidence} introduce a reinforcement learning fine-tuning framework that utilizes confidence signals to guide few-shot policy optimization. In contrast, for inference-time control, recent works such as DeepConf \citep{fu2025deepconf}, \citet{taubenfeld2025confidence}, and \citet{zhang2026confidence} employ confidence to dynamically modulate the decoding process, ensuring greater reasoning reliability without additional training.

In contrast to entropy-based approaches, our work adopts token-level predictive confidence as the uncertainty measure and directly leverages it to perform trajectory correction during decoding.
By intervening at low-confidence positions and refining the decoding trajectory accordingly, our method provides a simple and effective confidence-guided mechanism for improving reasoning reliability.

\subsection{Contrastive Decoding}

Contrastive Decoding (CD) improves generation quality by contrasting the output distributions of different models or different prompting conditions. This paradigm has been applied across both text-only and multimodal settings.

\textbf{Text-based CD methods} originated with \citet{li2023contrastive}, who contrast larger models against smaller amateur models to improve factuality. Subsequent work extended this paradigm: DoLA~\cite{chuang2023dola} contrasts logits between different transformer layers to mitigate hallucinations; ICD~\cite{wang2024mitigating} induces hallucinations through lightweight fine-tuning to create a "weak" model for contrast, penalizing fabricated information during decoding; and ACD~\cite{zhao2024adversarial} employs opposite prompt optimization to boost safety alignment through adversarial contrast.

\textbf{Multimodal CD methods} address hallucinations in vision-language models (VLMs). VCD~\cite{leng2024mitigating} contrasts distributions between original and visually perturbed inputs to reduce object hallucinations. While these methods~\cite{leng2024mitigating,zhuang2025vasparse} require auxiliary models, layer-wise computations, or multiple forward passes, our approach distinguishes itself by performing \emph{localized} contrastive intervention only at high-uncertainty tokens, significantly reducing computational overhead while maintaining or improving effectiveness across both textual and reasoning tasks.

\section{Appendix.}
\subsection{Case Study: Qualitative Comparison of Reasoning Traces}

We present a qualitative case study on an AIME 2025 combinatorics problem to illustrate how Confidence-Driven Contrastive Decoding (CCD) affects the reasoning process compared to standard chain-of-thought decoding.
Figure shows a side-by-side comparison between a Trivial CoT trace generated by the base decoding strategy and a CoT trace produced by CCD.

\begin{figure}[h]
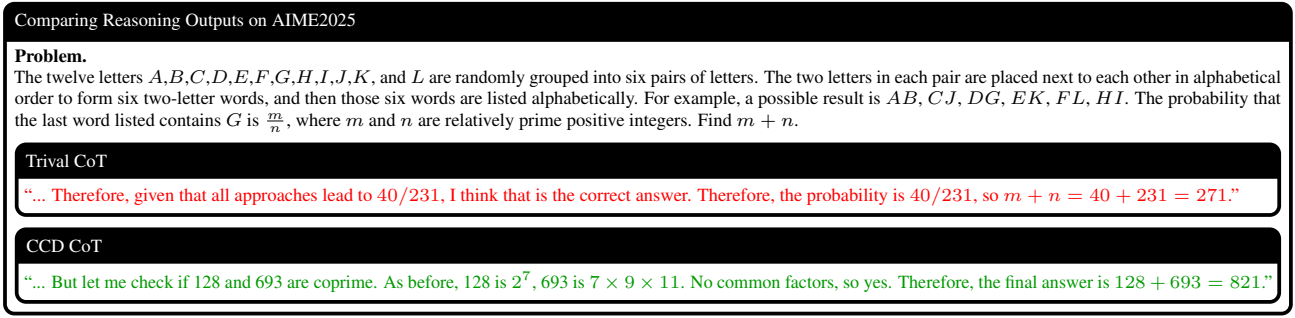

    \newtcolorbox{scopebox}[2]{%
        left=0pt,
        right=0pt,
        top=0pt,
        bottom=0pt,
        boxsep=3pt,
        colback=white,
        colframe=black,
        title={#1},
    }
    \centering
    \scriptsize
    \begin{scopebox}{Comparing Reasoning Outputs on AIME2025}{}
        \noindent\textbf{Problem.}

        The twelve letters $A$,$B$,$C$,$D$,$E$,$F$,$G$,$H$,$I$,$J$,$K$, and $L$ are randomly grouped into six pairs of letters.
        The two letters in each pair are placed next to each other in alphabetical order to form six two-letter words, and then those six words are listed alphabetically.
        For example, a possible result is $AB$, $CJ$, $DG$, $EK$, $FL$, $HI$.
        The probability that the last word listed contains $G$ is $\frac{m}{n}$, where $m$ and $n$ are relatively prime positive integers.
        Find $m+n$.

        \begin{scopebox}{Trival CoT}{}
            \wrong{``...
            Therefore, given that all approaches lead to $40/231$, I think that is the correct answer.
            Therefore, the probability is $40/231$, so $m+n = 40 + 231 = 271$.''}
        \end{scopebox}
        
        \begin{scopebox}{CCD CoT}{}
            \correct{``...
            But let me check if 128 and 693 are coprime.
            As before, 128 is $2^7$, 693 is $7 \times 9 \times 11$.
            No common factors, so yes.
            Therefore, the final answer is $128 + 693 = 821$.''}
        \end{scopebox}
    \end{scopebox}
    \caption{
        Case study on an AIME 2025 problem: comparison between Trivial CoT and CCD CoT on a combinatorics task.
    }
    \label{fig:case study}
\end{figure}

\subsection{Discussion: Stability vs.\ Exploration}
\paragraph{Stability versus exploration in reasoning.}
From a reasoning perspective, effective inference requires balancing two complementary forces:
exploiting strong semantic priors to maintain stability, and selectively relaxing them to allow exploration when uncertainty arises.
CCD primarily targets the stability side of this trade-off.
By suppressing spurious alternatives at low-confidence positions through contrastive subtraction, CCD reduces the risk of error amplification and maintains coherent reasoning trajectories.
This analysis helps explain why CCD yields consistent gains across diverse reasoning benchmarks despite its simplicity.

\subsection{Keywords Used for Chain-of-Thought (CoT) Frequency Analysis}

\begin{table}[ht]
    \centering
    \caption{Keywords Used for Chain-of-Thought (CoT) Frequency Analysis}
    \label{tab:keywords}
    \begin{tabularx}{\textwidth}{l X}
        \toprule
        \textbf{Category} & \textbf{Target Keywords} \\
        \midrule
        \textbf{Hesitation} & wait, Wait, WAIT, hmm, Hmm, HMM \\
        \addlinespace
        \textbf{Correction} & but, But, BUT, however, However, HOWEVER, wait,but, Wait,but, but wait, But wait, hold on, Hold on \\
        \addlinespace
        \textbf{Self-Correction} & let me double-check, Let me double-check, looking back, Looking back \\
        \addlinespace
        \textbf{Alternatives} & alternatively, Alternatively, ALTERNATIVELY, similarly, similarly \\
        \addlinespace
        \textbf{Verification} & seems solid, Seems solid, that's correct, but, That's correct, but, that seems right, That seems right \\
        \addlinespace
        \textbf{Markers} & SO \\
        \bottomrule
    \end{tabularx}
\end{table}

\end{document}